\documentclass[sigconf,nonacm]{acmart}

\setcopyright{none}
\acmDOI{}
\acmISBN{}
\usepackage{amsmath}
\usepackage{comment}
\usepackage{booktabs}
\usepackage{multirow}
\usepackage{makecell}
\usepackage{enumitem}
\usepackage{graphicx}
\usepackage{threeparttable}
\usepackage{tablefootnote}

\usepackage{CJKutf8}
\newcommand*{\Ja}[1]{%
  \begin{CJK}{UTF8}{ipxm}#1\end{CJK}}

\AtBeginDocument{%
  }

\begin{document}

\title{Forecasting Sales Delay through Time-aware Analysis of\\Sales Events in Textual Reports}


\author{Masaya Tsunokake}
\affiliation{%
  \institution{R\&D Group, Hitachi, Ltd.}
  \city{Tokyo}
  \country{Japan}}
\email{masaya.tsunokake.qu@hitachi.com}

\author{Yuta Koreeda}
\affiliation{%
  \institution{R\&D Group, Hitachi, Ltd.}
  \city{Tokyo}
  \country{Japan}}
\email{yuta.koreeda.pb@hitachi.com}
  
\author{Yuhei Niwa}
\affiliation{%
  \institution{R\&D Group, Hitachi, Ltd.}
  \city{Tokyo}
  \country{Japan}}
\email{yuhei.niwa.yh@hitachi.com}

\author{Takeshi Homma}
\affiliation{%
  \institution{R\&D Group, Hitachi, Ltd.}
  \city{Tokyo}
  \country{Japan}}
  
\author{Terufumi Morishita}
\affiliation{%
  \institution{R\&D Group, Hitachi, Ltd.}
  \city{Tokyo}
  \country{Japan}}
\email{terufumi.morishita.wp@hitachi.com}

\author{Teppei Inoue}
\affiliation{%
  \institution{R\&D Group, Hitachi, Ltd.}
  \city{Tokyo}
  \country{Japan}}
\email{teppei.inoue.wa@hitachi.com}

\author{Hiroaki Ozaki}
\affiliation{%
  \institution{Hitachi, Ltd.}
  \city{Tokyo}
  \country{Japan}}
\email{hiroaki.ozaki.yu@hitachi.com}

\author{Yasuhiro Sogawa}
\affiliation{%
  \institution{Hitachi, Ltd.}
  \city{Tokyo}
  \country{Japan}}
\email{yasuhiro.sogawa.tp@hitachi.com}
\renewcommand{\shortauthors}{Tsunokake et al.}

\begin{abstract}
Business-to-business (B2B) companies develop and deliver a one-of-a-kind product in response to a client's orders. To prepare resources for development beforehand, companies estimate dates when they will receive orders from clients in their sales process, called scheduled order dates.
However, dates when the companies actually receive orders can differ from the scheduled order dates. This difference increases costs in the development.
If the difference could be forecasted before receiving orders, companies could optimize their resource preparation.
Towards reducing the costs, we introduce \textbf{sales delay forecasting}, a novel task of forecasting delay days from the scheduled order dates during sales process where sales representatives (sales reps) interact with the clients to receive orders.
Sales delay forecasting is challenging because the sales process progress inherently involves uncertainty and depends on the client’s willingness to order.
To address this task, we propose \textbf{confidence features} which represent sales reps’ confidence as a proxy measure of the client’s willingness to order.
To represent it, our method extracts dates of future sales events anticipated by sales reps from textual reports.
To consider whether the confidence objectively looks reasonable, we also introduce \textbf{status features} which describe sales process status based on current date and past events.
In experiments using real sales data, our method achieved a 16\% lower mean absolute error compared with a feature-based baseline similar to a prior work. 
Moreover, our method outperformed non-feature-based baselines using large language models, indicating that it is crucial to explicitly create our features.
\end{abstract}


\maketitle

\begin{figure}[t]
 \centering
 \includegraphics[keepaspectratio, width=1.0\linewidth]{
    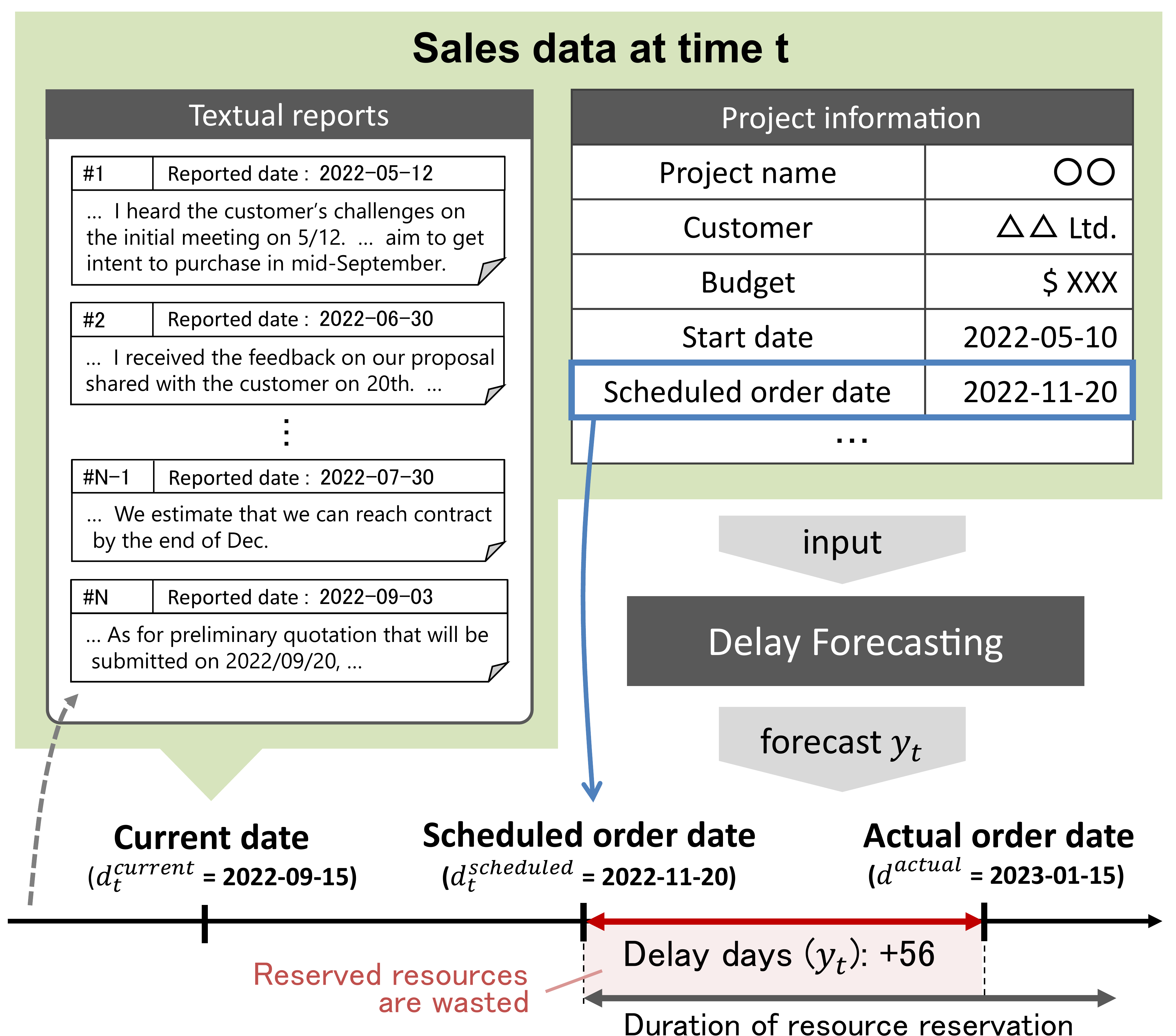
 }
 \caption{Sales delay forecasting. We address this task of forecasting delay days ($y_t$) from scheduled order date utilizing sales data at time $t$.\label{fig: task_structure}}
\end{figure}

\section{Introduction\label{sec: intro}}
Many business-to-business (B2B) companies employ a make-to-order strategy, where they develop and deliver a one-of-a-kind product in response to a client's order.
This process requires much time and resources.
Therefore, to start development quickly upon receiving the orders, B2B companies reserve development resources (e.g., workers, materials) in advance for the entire duration of development.
The duration of the reservation begins on a \textbf{scheduled order date}, which is the date that a B2B company estimates when the company will receive an order.
Since the company optimizes reservation plans to leverage the limited resources over multiple projects, the duration of reservation cannot be easily changed.
However, an \textbf{actual order date}, which is the date when the company actually receives the order, can differ from the scheduled order date.
This difference (i.e., $y_t$ = the actual order date - the scheduled order date) can incur additional costs. 
For example, if receiving an order is delayed ($y_t > 0$), the reserved resources are wasted for the duration of the delay, as shown in Figure \ref{fig: task_structure}.
On the other hand, if the receiving of an order is earlier ($y_t < 0$), it may incur additional expenses to immediately start development. 

B2B companies receive clients' orders through a \textbf{sales process} where sales representatives \textbf{(sales reps}) capture the clients' needs throughout continuous interaction. If $y_t$ could be forecasted during sales process, companies could optimize their resource reservation plan beforehand, leading to cost cutting.
Towards achieving this, we introduce a novel task called \textbf{sales delay forecasting} as shown in Figure \ref{fig: task_structure}.
This task aims to forecast delay days ($y_t$) from the scheduled order date before receiving an order on the actual order date, utilizing sales data accumulated until $t$.

Sales delay forecasting is more challenging than similar tasks, which aim to forecast delay days for other processes such as software development and construction, in prior works~\cite{hegazy2005_delayAnalysis,oliveros2005_delayAnalysis,ghimire2019_delayAnalysis_survey,son2019_tte,zadeh2022_tte}. 
Such a process steadily progresses by construction of artifacts (e.g., parts of software).
Thus, the lead time of the process can be roughly estimated based on the uncompleted parts of the artifact.
In contrast, a sales process is a consensus-building process where a sales rep reaches a mutual agreement with a client to receive an order.
This indicates that whether a sales process progresses inherently involves uncertainty, and depends on the client's willingness to order.
However, the client's willingness to order cannot be directly observed, which makes sales delay forecasting challenging.

\begin{figure*}[t]
 \centering
 \includegraphics[keepaspectratio, width=0.95\linewidth]{
    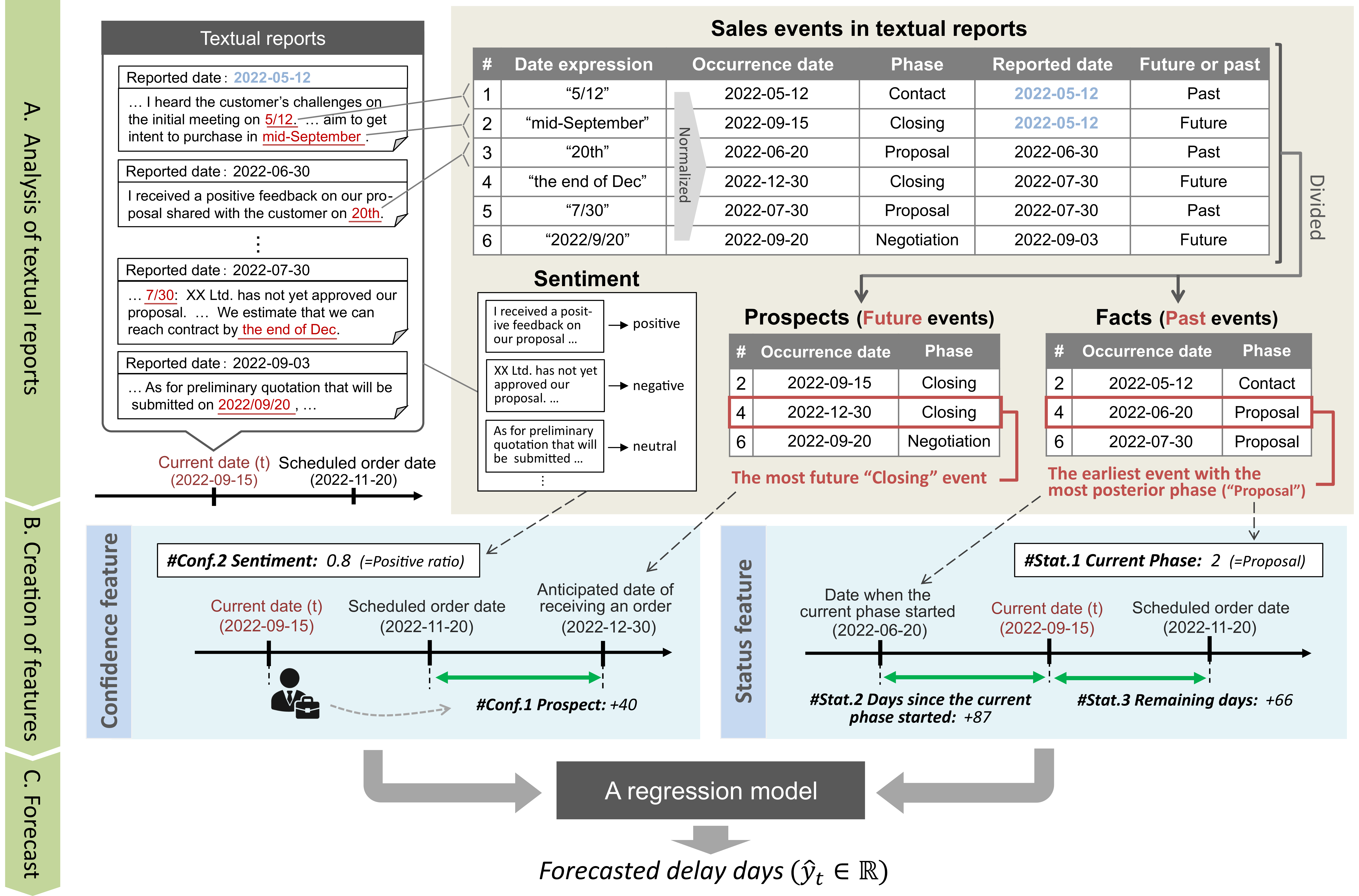
 }
 \caption{Overview of our method. Our method first extracts sales events from textual reports by identifying date expressions and keywords tied with sales phases, and then distinguishes whether the events refer to the future or the past. Example extracted events are shown in the top of this figure. Next, our method creates confidence and status features based on the extracted events. Confidence features represent sales reps' confidence and are created through utilizing the extracted future events (\#Conf.1) and sentiments detected from textual reports (\#Conf.2). Status features objectively describe sales process status. Status features are created through utilizing remaining days until the scheduled order date (\#Stat.3) and the extracted past events that determine the current phase (\#Stat.1) and days since the current phase started (\#Stat.2).
 \label{fig: proposed_method_overview}}
\end{figure*}

To address this, we propose \textbf{confidence features} which represent sales reps' confidence as a proxy measure of a client's willingness to order, as shown in Figure \ref{fig: proposed_method_overview}.
Through conversations with a client, a sales rep develops future prospects about the client's decisions based on the client's reactions and writes the prospects into textual reports.
Our method extracts such prospects by identifying dates of anticipated future events written in textual reports to represent sales reps' confidence.

On the other hand, sales reps' confidence is potentially unreliable.
Intuitively, when the scheduled order date is closing but the sales process is still in its initial phase (e.g., just contacting), $y_t$ becomes large even if the sales rep is highly confident in the process.
To avoid depending solely on potentially unreliable features, we also introduce \textbf{status features} to consider whether the confidence objectively looks reasonable based on sales process status such as remaining days until the scheduled order date and a current phase which is determined from pre-defined phases in Figure \ref{fig: sales_phase}. 

Using real sales data, we evaluated our method against the following baselines: 1) a feature-based baseline similar to a prior work in software development~\cite{zadeh2022_tte}, 2) non-feature-based baselines in which language models (e.g., RoBERTa, Qwen3) directly forecast $y_t$ based on textual reports.
Our method achieved a 16\% lower mean absolute error in forecasting $y_t$ compared with the feature-based baseline, indicating the superiority of our feature.
Our method also outperformed non-feature-based baselines using the same textual reports, indicating that it is crucial to explicitly create our features from the reports.
Through an ablation study, we confirmed that both confidence and status features contribute to forecasting performance. 
We also confirmed that the errors (i.e., $|y_t - \hat{y}_t|$) of our method notably decreased as the actual order date approaches (i.e., time $t$ increases). 
This result suggests that our features capture the prospects and status updated as time $t$ increases, and serve as a better indicator of forecasting.

Our contributions are as follows:
\begin{itemize}
    \item Towards reducing costs in B2B business, we introduce the novel task of \textbf{sales delay forecasting}, which aims to forecast delay days ($y_t$) from a scheduled order date during a sales process.
    \item To address sales delay forecasting, we propose a forecasting method based on novel features that represent sales reps' confidence and sales process status.
    \item In experiments using real sales data, we show that our method outperformed a feature-based baseline based on a prior work~\cite{zadeh2022_tte} and non-feature-based baselines wherein language models directly forecast $y_t$. The results indicate the effectiveness of our feature-based approach and the design.
    \item Further analysis suggests that our features capture the prospects and status updated as time $t$ increases, and serve as a better indicator of forecasting.
\end{itemize}

\section{Task Description\label{sec: task}}
The objective of sales delay forecasting is to forecast delay days from a scheduled order date before receiving an order from a client on the actual order date.
Figure \ref{fig: task_structure} illustrates this task. Sales data contain sales project information (e.g., project name and budget) and textual reports written by sales reps. Each report is attached with a \textbf{reported date}. 
The sales data is periodically updated.

We formulate sales delay forecasting as a regression problem as in similar tasks in prior works~\cite{son2019_tte,zadeh2022_tte}. In our setting, objective and dependent variables change depending on time $t$. Thus, this task is formulated as follows:
\begin{equation}
d^{\text{actual}} - d^{\text{scheduled}}_t := y_t = f(x^p, x_t, d^{\text{scheduled}}_t)
\end{equation}
where $t$ is the time of each forecasting point, $y_t$ is the delay days from the scheduled order date ($d^{\text{scheduled}}_t$) at time $t$, $d^{\text{actual}}$ is the actual order date, $x^p$ is the project information, and $x_t$ is information extracted from textual reports at time $t$.
Note that $y_t$ can be a negative value if the sales rep receives the order before $d^{\text{scheduled}}_t$.

\section{Proposed Method\label{sec: method}}
\begin{figure}[t]
 \centering
 \includegraphics[keepaspectratio, width=0.95\linewidth]{
    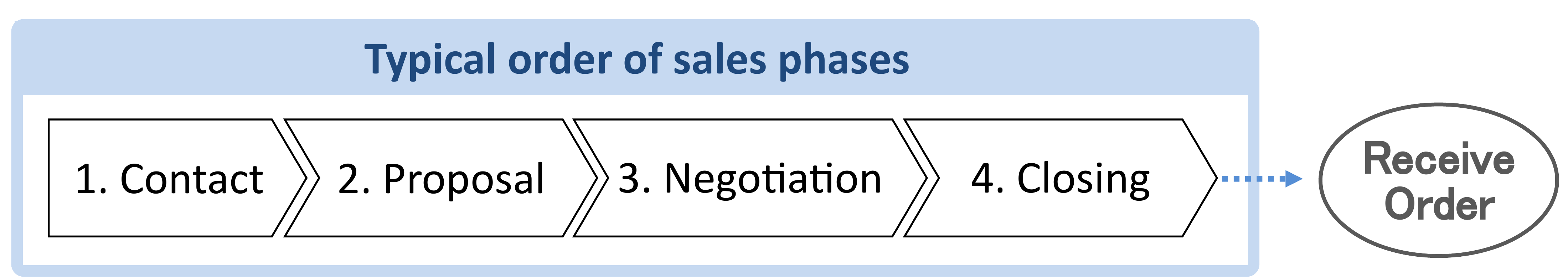
 }
 \caption{Sales phases. Generally, sales process finishes via sales events related to these phases.\label{fig: sales_phase}}
\end{figure}
\subsection{Overview \label{subsec: overview_method}}
Our method creates confidence features which represent sales reps’ confidence as a proxy measure of the client’s willingness to order.
On the other hand, sales reps’ confidence is potentially unreliable due to sales reps' biases.
To avoid depending solely on potentially unreliable features, our method also creates status features to consider whether the confidence objectively looks reasonable based on sales process status.

As for confidence features, our method utilizes sales reps' prospects about the client’s decisions, particularly the anticipated date of receiving orders.
Intuitively, if an anticipated date of receiving an order is substantially later than the scheduled order date, the sales rep's confidence is low, indicating a high likelihood of larger $y_t$.
To quantify the gap between the anticipated date of receiving orders and the scheduled order date, our method extracts dates of anticipated future events related to the order from textual reports.
Our method also detects sentiments in textual reports to capture the sales rep's confidence more directly.

As for status features, our method utilizes remaining days until the scheduled order date and a current phase of a sales process.
To determine the current phase, our method extracts facts of sales events in the past (i.e., past events) from textual reports while associating each past event to pre-defined phases in Figure \ref{fig: sales_phase}.
Our method regards the most posterior phase in past events as the current phase.

Both confidence and status features are based on sales events written in textual reports.
Hence, our method consists of three steps: (A) analysis of textual reports, (B) creation of features, and (C) forecast. 
Figure \ref{fig: proposed_method_overview} illustrates three steps.

\subsection{Analysis of textual reports\label{subsec: method_step1}}
\begin{table*}[t]
\small
\caption{Our features. Confidence features represent sales reps' confidence. Status features describe sales process status to consider whether the confidence objectively looks reasonable.\label{tab:features}} 
\centering
\setlength{\tabcolsep}{6pt}
{\renewcommand\arraystretch{1.0}
\begin{tabular}{cccl}
\toprule
Feature Type & Name & Description & \multicolumn{1}{c}{How to compute} \\
\midrule
\multirow{2}{*}{\makecell[tc]{Confidence}} & \makecell[tl]{\#Conf.1 Prospect} & \makecell[tl]{Days between date of most future event in future events with\\ ``Closing'' phase ($d^{\text{closing}}_t$) and scheduled order date ($d^{\text{scheduled}}_t$)} & \makecell[tl]{$d^{\text{scheduled}}_t - d^{\text{closing}}_t$} \\ \cmidrule(lr){2-4}
& \makecell[tl]{\#Conf.2 Sentiment} & \makecell[tl]{Ratio of positive sentences in textual reports} & $\frac{\text{number of sentences with positive sentiment}}{\text{number of all sentences}}$ \\ \midrule
\multirow{3}{*}{\makecell[tc]{Status}} & \makecell[tl]{\#Stat.1 Current phase} & \makecell[tl]{An ordinal number of the most posterior phase in past events} & \makecell[tl]{See ordinal numbers in Figure 3} \\ \cmidrule(lr){2-4}
& \makecell[tl]{\#Stat.2 Days since the cur-\\\quad\quad\quad rent phase started} & \makecell[tl]{Days between current date ($d^{\text{current}}_t$) and the earliest date in\\past events with current phase ($d^{\text{start}}_t$)} & $d^{\text{current}}_t - d^{\text{start}}_t$ \\ \cmidrule(lr){2-4}
& \makecell[tl]{\#Stat.3 Remaining days} & \makecell[tl]{Days between current date ($d^{\text{current}}_t$) and scheduled order date\\($d^{\text{scheduled}}_t$)} & $d^{\text{scheduled}}_t - d^{\text{current}}_t$ \\
\bottomrule
\end{tabular}
}
\end{table*}

This first step extracts sales events and their occurrence dates by identifying co-occurrences of a mention to a sales event and a date expression in textual reports.

To identify date expressions, our method uses temporal expression extractor~\cite{verhagen2007_tempExp,verhagen2010_tempExp,uzzaman2013_tempExp}.
There are various types of temporal expressions in text.
Our method only extracts temporal expressions whose types are ``DATE'' in the TimeML scheme~\cite{pustejovsky2003_tempExp} as date expressions and normalizes them, as shown in the top of Figure \ref{fig: proposed_method_overview}.
When normalizing, there can be a date expression whose year, month, or day is missing.
A missing year, month or day is complemented with the reported date of the textual report where the date expression is written.
In the following steps, our method creates a sales event table, as shown in Figure \ref{fig: proposed_method_overview}, based on these normalized date expressions (\textbf{occurrence dates}).

Next, our method identifies mentions to a sales event in the sentence from which the date expressions were identified.
Generally, a sales process has some phases towards receiving orders.
This paper defines four \textbf{sales phases}: ``Contact,'' ``Proposal,'' ``Negotiation,'' and ``Closing.''
Figure \ref{fig: sales_phase} illustrates the phases. 
We focus on events that are important for each phase since irrelevant events are also written in textual reports.
Our method associates each date expression with one of the sales phases by searching for the pre-defined keywords about important sales events \footnote{Keywords are designed on the basis of the domain and language of textual reports.}.
Date expressions that cannot be associated with the keywords are removed in this step.

Finally, to distinguish whether each sales event is a sales rep's prospect, our method identifies whether the occurrence date refers to the future or the past. 
Specifically, if the occurrence date refers to a date after the reported date, our method regards the corresponding event as a sales rep's prospect.
Otherwise, the event is regarded as a past event. The top of Figure \ref{fig: proposed_method_overview} shows sales events with all information described in this section.

\subsection{Creation of features\label{subsec: method_step2}}
This step creates confidence and status features, listed in Table \ref{tab:features}, utilizing extracted sales events and other information.

\subsubsection{Confidence features\label{subsubsec: subj_features}}
These features represent sales reps' confidence and include \textbf{\#Conf.1: Prospect} and \textbf{\#Conf.2: Sentiment}.

For \#Conf.1, we quantify a sales rep's confidence by measuring the difference between the scheduled order date ($d^{\text{scheduled}}_t$) and the date of receiving an order anticipated by the sales rep.
To determine the latter date, our method uses future events associated with ``Closing'' phase which typically occur just before receiving order.
The occurrence date of the most future event in the future events with ``Closing'' phase ($d^{\text{closing}}_t$) has the highest likelihood of the date of receiving an order.
Thus, our method computes the number of days between $d^{\text{closing}}_t$ and $d^{\text{scheduled}}_t$ as \#Conf.1 .

If sales reps are more confident, positive sentiments should appear more frequently.
Therefore, our method also computes the ratio of positive sentences in textual reports as \#Conf.2.
We utilize sentiment detection techniques studied in the field of natural language processing~\cite{socher2013_sentiment,pang2004_sentiment,rosenthal2017_sentiment}.

\subsubsection{Status features\label{subsubsec: obj_features}}
These features are designed to consider whether sales reps' confidence objectively looks reasonable. This feature set includes \textbf{\#Stat.1: Current phase}, \textbf{\#Stat.2: Days since the current phase started}, and \textbf{\#Stat.3: Remaining days}.

Our method obtains the most posterior phase in the past events and utilizes the ordinal number of the obtained phase as \#Stat.1 .
By utilizing the most posterior phase, we can approximately and objectively quantify how close a current sales process is to receiving an order.

Although \#Stat.1 characterizes the sales process status, it is coarse-grained.
To describe the sales process status at finer granularity, our method also computes the number of days since the current phase started as \#Stat.2 .
To compute it, the date of the earliest past event associated with the current phase ($d^{\text{start}}_t$) is regarded as the starting date, as shown in Figure \ref{fig: proposed_method_overview}.

We quantify not only the current phase but also the remaining days until the scheduled order date as \#Stat.3 . Concretely, our method computes the number of days between $d^{\text{scheduled}}_t$ and a current date ($d^{\text{current}}_t$).

\section{Experiment\label{sec: experiment}}
By comparing our method, another feature-based method based on a prior work~\cite{zadeh2022_tte}, and non-feature-based baselines using language models, we verify the effectiveness of our method that explicitly creates confidence and status features from textual reports.

\subsection{Experimental Setup\label{subsec: setup}}
\subsubsection{Data\label{subsubsec: data}}
 We used real sales data of a Japanese company. It consists of closed sales projects. To evaluate throughout the whole sales process, we set multiple forecasting points (i.e., time $t$) for each sales project. As the sales process status is updated, textual reports increase. Therefore, we set $t$ every time reports were added. At each $t$, we created features using the reports accumulated up to $t$.
 Since our target projects are completely independent of each other, the sales projects in the data were randomly split into training (train), development (dev), and test sets.
 Note that each set contains multiple instances of time $t$ for a single sales project.
 To simplify the comparison of different selections of features, we only experimented with time $t$ at which all features could be created. The train, dev, and test sets have 1,242, 245, and 315 instances of time $t$, respectively.

\subsubsection{Baselines\label{subsubsec: baselines}}
We introduce the following baselines.
\paragraph{\textbf{Trivial baselines}} We introduce two trivial baselines: 1) \textbf{Zero} that constantly predicts that the sales process would not be delayed and thus outputs 0 as $y_t$, and  2) \textbf{Mean} that constantly outputs the mean of $y_t$ in the train set.
\paragraph{\textbf{ProjectInfo}} We introduce a feature-based baseline using only project information that represent similar information as in a prior work~\cite{zadeh2022_tte}. The features consist of an initial scheduled duration (i.e., days between scheduled order date and date when a sales project is launched) and a budget which means a planned revenue of the sales project.
\paragraph{\textbf{Finetuned-LM}} To verify the effectiveness of explicit feature creation from textual reports, we set non-feature-based baselines that directly forecast $y_t$ from same textual reports by using a fine-tuned language model (LM) without explicitly creating features. We verified three LMs which are common choices and successful in many applications:  RoBERTa~\cite{liu2019robertarobustlyoptimizedbert_roberta}, LUKE~\cite{yamada-etal-2020-luke}, and BigBird~\cite{zaheer2020big}.
Such LMs have acquired the capability of understanding textual semantics through pre-training on large textual corpora.
We fine-tune each LM with an additional linear layer and MSE loss. Input text was created by concatenating the three latest textual reports and remaining days until scheduled order dates (i.e., \#Stat.3).

\paragraph{\textbf{Fewshot-LLM}} As another non-feature-based baseline, we verify few-shot approach by large language model (LLM). Recent LLMs exhibit remarkable performances for unseen difficult tasks given few-shot samples in the prompt~\cite{brown2020language,dong2024surveyincontextlearning,openai2024gpt4technicalreport}. Due to the confidentiality of real sales data, we cannot use closed-models.
Instead, we use Qwen3-32B~\cite{qwen3} that has competitive performance with top-tier LLMs such as OpenAI-o1~\cite{openai2024o1} and DeepSeek-R1~\cite{deepseekai2025deepseekr1} and is compatible with Japanese. We also employ Qwen3-8B~\cite{qwen3} with distinct model size.
Same as Finetuned-LM, input text was created by concatenating the three latest textual reports and \#Stat.3 . We also provide 5-shot samples consisting of the input text and $y_t$.

\subsubsection{Hyperparameters \label{subsubsec: hyperparameters}}
Since the best machine learning method (ML) varies on features, we set the type of ML as one of the hyperparameters to compare our method and ProjectInfo on their best settings. As candidates of MLs, we choose common MLs: SVM~\cite{vapnik1995nature_SVM_foundation} with linear or RBF kernel, and Random Forest~\cite{breiman2001random}. We selected the ML and hyperparameters\footnote{For SVM, we tested 0.0001, 0.001, 0.1, 1, and 10 as regularization parameters and 1,000 as the maximum number of iterations within solvers. For Random Forest, we tested unlimited, 5, and 10 as the maximum depth and 2, 10 as the minimum number of samples required to split internal nodes. We also tested 1, 3 as the minimum number of samples at a leaf node and 100, 1,000 as the number of weak learners. We also tested setting with and without standardization of features.} exhibiting the lowest MAE in the dev set and evaluated the selected ML trained with the selected hyperparameters. We report the results averaged over five random seeds.

For Finetuned-LM baseline, we used Japanese RoBERTa, LUKE, and BigBird\footnote{HuggingFace tags: ku-nlp/roberta-large-japanese-char-wwm (RoBERTa), studio-ousia/luke-japanese-large-lite (LUKE), nlp-waseda/bigbird-base-japanese (BigBird)}.
Each LM was fine-tuned on the train set up to 10 epochs. The training was terminated if the minimum loss on the dev set could not be updated within 3 epochs. The optimization function was AdamW~\cite{loshchilov2017_adam}. We verified the performances on five random seeds. For each random seed, we selected hyperparameters\footnote{The learning rate candidates were 1.0e-4, 5.0e-5, 1.0e-5, 5.0e-6, and 1.0e-6. The maximum length of input tokens was 512 for RoBERTa and LUKE, and 4096 for BigBird. The batch sizes for RoBERTa and LUKE were 32, and that of BigBird was 4.} exhibiting the lowest MAE on the dev set and evaluated the selected model on the test set.

For Fewshot-LLM baseline, we created 5-shot samples that are randomly selected from forecasting points ($t$) in the train set.
We enable Qwen3's \textit{thinking mode}~\cite{qwen3} which explicitly instructs the model to output intermediate reasoning towards final answers. 
As for generation parameters, we followed best practices in HuggingFace page\footnote{e.g., \url{https://huggingface.co/Qwen/Qwen3-32B}}.
We report the results averaged over five random seeds which affect the 5-shot sample selection and the generation probability.

\subsubsection{Evaluation\label{subsubsec: metric}}
We evaluated the errors in forecasts by using the mean absolute error (MAE) and root mean squared error (RMSE). RMSE squares the error so that it incurs a larger penalty compared with MAE when the model makes a large mistake.

\subsection{Implementation}
In our method, we split the textual reports into sentences using the \verb|ja_ginza| model via GINZA\footnote{\url{https://github.com/megagonlabs/ginza}} and spaCy\footnote{\url{https://spacy.io/}}. Date expressions in each sentence were detected and normalized using ja-timex\footnote{\url{https://github.com/yagays/ja-timex}}, which is a rule-based parser based on the Japanese TimeML-like scheme~\cite{asahara2014_tempExp}.
To associate date expressions with the sales phases, we search for the keywords, listed in Table \ref{tab:keywords}, in the sentences.
When associating date expressions with the pre-defined sales phases shown in Figure \ref{fig: sales_phase}, we used keywords in Table \ref{tab:keywords}.
Sentiment was detected by Asari\footnote{\url{https://github.com/Hironsan/asari}}.
MLs were implemented by scikit-learn~\cite{scikit_learn}. We use HuggingFace to implement Finetuned-LM and Fewshot-LLM. The LMs were fine-tuned using NVIDIA A100 GPUs.

\subsection{Results \label{subsec: main_result}}
\begin{table}[t]
\centering
\small
\begin{threeparttable}
\caption{Example keywords tied with sales phases\label{tab:keywords}}
\begin{tabular}{ll}
\toprule
\multicolumn{1}{c}{Event type} & \multicolumn{1}{c}{Keywords\tnote{a}} \\
\midrule
Contact & \makecell[tl]{\Ja{キックオフ}(initial meeting), \Ja{ヒアリング}(hearing),\\ \Ja{引き合い}(inquiry), \Ja{引合}(inquiry)} \\ \midrule
Proposal & \makecell[tl]{\Ja{提案}(proposal), rfp(request for proposal)} \\ \midrule
Negotiation & \makecell[tl]{\Ja{見積}(quotation), \Ja{要件定義}(requirements definition)} \\ \midrule
Closing & \makecell[tl]{\Ja{内示}(receiving intent to purchase), \Ja{発注}(order)\\\Ja{受注}(receiving order), \Ja{検収}(receipt inspection)\\\Ja{契約}(contract), \Ja{注文}(purchase order)}\\
\bottomrule
\end{tabular}
\begin{tablenotes}
\item[a] English translations follow original Japanese keywords.
\end{tablenotes}
\end{threeparttable}
\end{table}
\begin{table}[t]
\centering
\caption{MAE and RMSE of delay days ($y_t$) forecasted by each method. Our method achieved lower MAE than all baselines.\label{tab:exp_main_result}}
\setlength{\tabcolsep}{5pt}
{\renewcommand\arraystretch{1.0}
\begin{tabular}{llrr}
\toprule
\multicolumn{2}{c}{Method} & \multicolumn{1}{c}{MAE} & \multicolumn{1}{c}{RMSE} \\ \midrule
\multicolumn{2}{@{}l}{Baselines} & & \\
& Mean & 91.06 & \textbf{109.95} \\
& Zero & 96.29 & 142.66 \\
& ProjectInfo (similar to~\cite{zadeh2022_tte}) & 84.06 & 123.41 \\
& Finetuned-LM w/ RoBERTa & 83.72 & 122.86\\
& Finetuned-LM w/ LUKE & 83.73 & 122.94\\
& Finetuned-LM w/ BigBird & 92.41 & \underline{110.10}\\
& Fewshot-LLM w/ Qwen3-8B & \underline{76.88} & 127.80\\
& Fewshot-LLM w/ Qwen3-32B & \underline{76.10} & 127.48\\
\multicolumn{2}{@{}l}{\textbf{Our Method}} & \textbf{70.80} & \underline{112.82} \\
\midrule
\multicolumn{2}{@{}l}{Ablation Study} & & \\
& w/o \#Conf.1 Prospect & $+44.17$ & $+46.47$ \\
& w/o \#Conf.2 Sentiment & $+23.20$ & $+40.51$ \\
& \makecell[tl]{w/o \#Stat.1 Current phase} & $+28.14$ & $+42.93$ \\
& \makecell[tl]{w/o \#Stat.2 Days since the\\\quad\quad\quad\quad\quad current phase started} & $-1.19$ & $+1.20$ \\
& \makecell[tl]{w/o \#Stat.3 Remaining days} & $+13.75$ & $+2.86$ \\
\bottomrule
\end{tabular}
}
\end{table}
Table \ref{tab:exp_main_result} presents the results\footnote{The selected ML for both ProjectInfo and our method was SVM with RBF kernel.}.
Our method achieved the lowest MAE among all methods. 
In particular, our method achieved 16\% lower MAE than ProjectInfo baseline based on the prior work~\cite{zadeh2022_tte}, which demonstrates that it is effective to consider not project information but rather our confidence and status features.

\begin{figure}[t]
 \centering
 \includegraphics[keepaspectratio, width=0.945\linewidth]{
    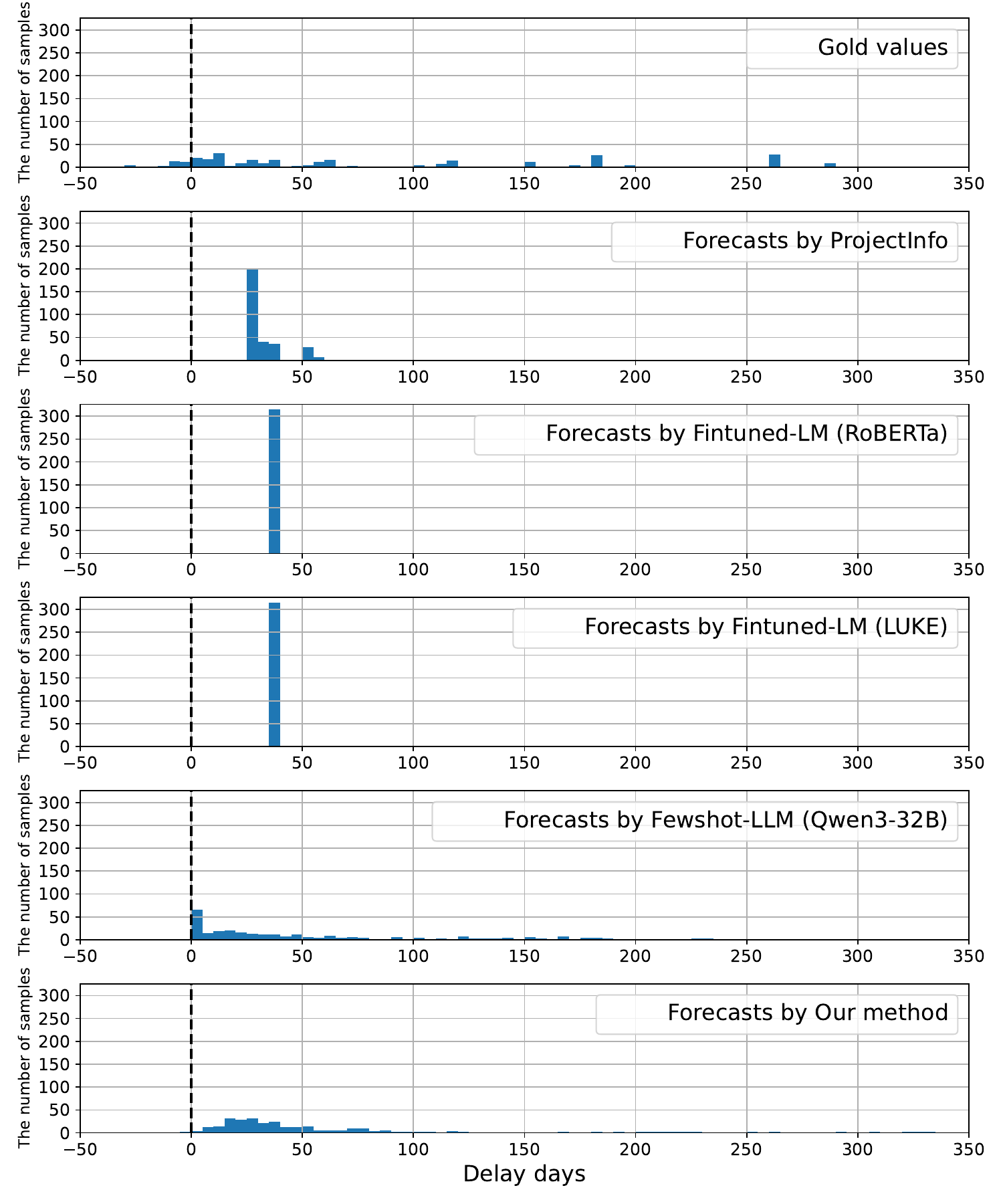
 }
 \caption{Histogram of gold delay days ($y_t$) on the test set and delay days ($\hat{y}_t$) forecasted by each method. Although the gold delay days in the top of this figure were widely distributed, Finetuned-LM forecasted a constant value.\label{fig: histogram_pred_value}}
\end{figure}

Our method also exhibits lower MAE than all Finetuned-LM baselines.
Figure \ref{fig: histogram_pred_value} shows a histogram of gold delay days on the test set and the delay days ($y_t$) forecasted by different methods\footnote{Due to the space limitations, we exclude the Zero and Mean baselines that definitively output constant values and focus on the forecasted delay days by the best finetuned-LM and the best Fewshot-LLM.}.
Finetuned-LMs such as LUKE and BigBird forecasted certain values, although the gold values were widely distributed.
This suggests that sales delay forecasting is too noisy to capture solely from text without inductive biases introduced by features, and optimized LMs fell back to predicting certain values. 
In comparison, our method can forecast a broad range of values while achieving the lowest MAE among all methods.
This indicates that it is crucial to explicitly design our features from textual reports.

Both MAE and RMSE of all Fewshot-LLM baselines are inferior to our method, emphasizing the superiority of our method.
As shown in Figure \ref{fig: histogram_pred_value}, the forecasts of Fewshot-LLM are slightly biased to $\hat{y}=0$ although they are widely spread compared to other baselines.

Mean baseline and Finetuned-LM w/ BigBird exhibited lower RMSEs than our method.
This is attributed to the forecasting tendency of these methods. 
Mean baseline constantly outputs the mean of $y_t$ on the train set. 
As shown in Figure \ref{fig: histogram_pred_value}, Finetuned-LM w/ BigBird also forecasted a certain value.
This characteristic of passive forecasting suppresses large errors.
Since RMSE imposes a larger penalty when the model makes a large mistake, these methods tend to be highly rated. Our method achieved competitive RMSE against these methods while forecasting various values.

\section{Analysis\label{sec: analysis}}

\subsection{Ablation study\label{subsec: ablation_study}}
To clarify which features are important in forecasting performance, we examined the change in the performance when each feature was excluded. The hyperparameters for this ablation study were selected using the same procedure as described in Section \ref{subsec: setup}.

Table \ref{tab:exp_main_result} presents the results. Each row with ``w/o feature name'' shows the change in the performance when the corresponding feature was excluded. When some confidence and status features were excluded, MAE and RMSE substantially got worse. This suggests that incorporating both confidence and status features is important for sales delay forecasting.

\subsection{Performances with different time $t$\label{subsec: eval_on_some_timing}}
\begin{figure}[t]
 \centering
 \includegraphics[keepaspectratio, width=0.97\linewidth]{
    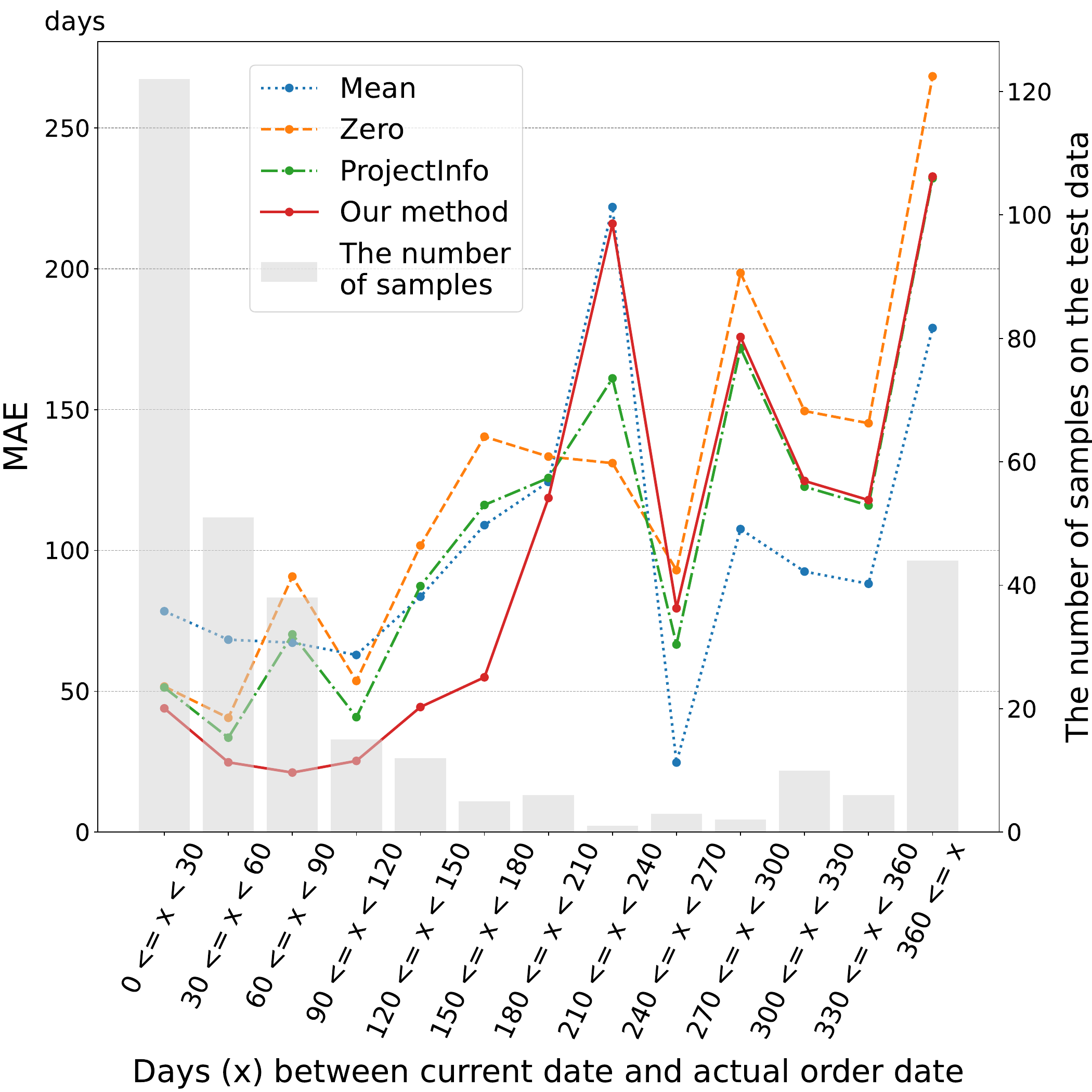
 }
 \caption{MAEs with different time $t$. The horizontal axis indicates days ($x$) between current date $d^{\text{current}}_t$ and actual order date $d^{\text{actual}}$ (i.e., $x$ = $d^{\text{actual}}$-$d^{\text{current}}_t$). The position to the left indicates that more time $t$ has passed. The vertical axis indicates MAE on each evaluation interval of $x$.\label{fig: eval_on_some_timing}}
\end{figure}
Intuitively, when a sales process has just started, it is difficult to forecast how smoothly the sales process will progress in the future.
In contrast, as a sales project gets updated in the course of the sales process, the sales reps' prospects become more accurate. Forecasting $y_t$ may become easier by capturing the updated prospect (confidence features) and status (status features).
Therefore, we assumed that our method can more accurately forecast as the actual order date approaches (i.e., time $t$ increases).

To represent how close sales project at time $t$ is to the actual order date, we set discrete intervals based on days between the current date ($d^{\text{current}}_t$) and the actual order date ($d^{\text{actual}}$).
We selected one random seed and investigated MAE in each interval on test set. 
Figure \ref{fig: eval_on_some_timing} shows the result.
The position to the left in Figure \ref{fig: eval_on_some_timing} indicates that more time has passed in the sales process.
In other words, MAE of our method largely decreased as time $t$ increases in the sales process, and the rate of decrease (i.e., $\frac{MAE_{t2} - MAE_{t1}}{t2 - t1}$) was higher than the baselines.
This result suggests that our features capture the prospects and status updated as time $t$ increases, and serve as better indicators of forecasting.

The MAEs of all methods except Mean baseline tended to decrease as time $t$ increases.
This shows that it was more difficult to forecast delay days on the basis of information obtained in the initial period. 
Mean baseline exhibited a relatively stable performance at the early time in the right side in Figure \ref{fig: eval_on_some_timing}. This is presumably because the Mean baseline outputs a certain value instead of relying on unsettled information obtained early.

\subsection{Impact of the thinking by LLMs\label{subsec: analysis_llm}}
In Table \ref{tab:exp_main_result}, Fewshot-LLM by Qwen3-8B and 32B invoked their thinking mode which explicitly instructs the model to output intermediate reasoning towards final answers. Without thinking mode, MAEs of Qwen3-8B and 32B got increased by +8.13 and +8.20, indicating that thinking mode is beneficial. Observing the thinking processes, the LLMs were trying to forecast by arithmetic operations using date expressions of sales events in textual reports. This emphasizes the importance of utilizing date expressions as in our method. On the other hand, our method outperformed Fewshot-LLM baselines with such thinking, indicating that our feature design is effective. Moreover, thinking by LLMs requires substantially more inference time than our feature-based method.

\section{Related work\label{sec: related_work}}
In the construction industry, delays frequently occur and might cause disputes between contractors and clients. Since the causes of the delays will determine the assignment of responsibilities between them, the construction is strictly managed with fine-grained milestone tables that are composed of many tasks and their task duration.
Previous studies propose various methodologies for identifying the cause of an overall delay from a delay of each task~\cite{hegazy2005_delayAnalysis,oliveros2005_delayAnalysis,yang2009_delayAnalysis_survey,fagiar2023_delayAnalysis,ghimire2019_delayAnalysis_survey}.
While many methodologies are based on the manual analysis by humans~\cite{yang2009_delayAnalysis_survey,fagiar2023_delayAnalysis,ghimire2019_delayAnalysis_survey}, Hegazy and Zhang~\cite{hegazy2005_delayAnalysis} proposed a computational method that estimates the total delay days of the whole construction process by aggregating the delay information of each task. Oliveros and Fayek~\cite{oliveros2005_delayAnalysis} also propose a computational method using fuzzy logic.
However, it is difficult to create such fine-grained milestone tables in consensus-building process such as sales process.

Some studies propose methods for forecasting delay days without relying on delay information of constituting tasks~\cite{son2019_tte,zadeh2022_tte}.
Zadeh and Kashef~\cite{zadeh2022_tte} verified the estimation of delay days on the basis of the initial project budgets and the initial project duration.
In this paper, we verified ProjectInfo using similar information that can be obtained on sales projects and confirmed that our method outperformed ProjectInfo. 

\section{Conclusion}
We introduced the novel task of sales delay forecasting. To address this task, we proposed confidence features which represent sales reps’ confidence as a proxy measure of the client’s willingness to order. We also proposed status features based on sales process status.
On real sales data, our method achieved lower MAE by 16\% compared with the feature-based baseline similar to a prior work~\cite{zadeh2022_tte}. Comparison with the non-feature-based baselines revealed the effectiveness of our feature-based approach. Further analysis suggests that our features capture the prospects and status updated as time $t$ increases, and serve as a better indicator of forecasting.

\begin{acks}
We would like to thank Kana Ozaki, Atsuki Yamaguchi, Gaku Morio, Koichi Nagatsuka, Yuichi Sasazawa, Kouki Takeshita, and Yumi Hamazono for their valuable feedback and Masaaki Shimizu for maintenance and management of large computational resources.
\end{acks}

\section*{GenAI Usage Disclosure}
Generative AI models were utilized partially to modify some sentences and figures.

\bibliographystyle{ACM-Reference-Format}
\bibliography{my_main/refs}

@article{hegazy2005_delayAnalysis,
  title={Daily windows delay analysis},
  author={Hegazy, Tarek and Zhang, Kehui},
  journal={Journal of Construction Engineering and Management},
  volume={131},
  number={5},
  pages={505--512},
  year={2005}
}

@article{oliveros2005_delayAnalysis,
  title={Fuzzy logic approach for activity delay analysis and schedule updating},
  author={Oliveros, Adriana V Ord{\'o}{\~n}ez and Fayek, Aminah Robinson},
  journal={Journal of Construction Engineering and Management},
  volume={131},
  number={1},
  pages={42--51},
  year={2005}
}

@article{yang2009_delayAnalysis_survey,
  title={Review of delay analysis methods: A process-based comparison},
  author={Yang, Jyh-Bin and Kao, Chih-Kuei},
  journal={The Open Construction \& Building Technology Journal},
  volume={3},
  pages={81--89},
  year={2009}
}

@article{fagiar2023_delayAnalysis,
  title={Simulation-Assisted Project Data Integration for Development and Analysis of As-Built Schedules},
  author={Fagiar, Muaz and Mohamed, Yasser and AbouRizk, Simaan},
  journal={Buildings},
  volume={13},
  number={4},
  pages={974},
  year={2023}
}

@article{ghimire2019_delayAnalysis_survey,
  title={Comparative study of prospective delay analysis techniques ({DATs})},
  author={Ghimire, Suroj and Mishra, Anjay Kumar},
  journal={Saudi Journal of Civil Engineering},
  volume={3},
  number={5},
  pages={84--98},
  year={2019}
}

@article{son2019_tte,
  title={Using text mining to estimate schedule delay risk of 13 offshore oil and gas {EPC} case studies during the bidding process},
  author={Son, Byung-Yun and Lee, Eul-Bum},
  journal={Energies},
  volume={12},
  number={10},
  pages={1956},
  year={2019}
}

@article{zadeh2022_tte,
  title={The Impact of {IT} Projects Complexity on Cost Overruns and Schedule Delays},
  author={Zadeh, Mahta Taghi and Kashef, Rasha},
 journal={arXiv preprint arXiv:2203.00430},
  year={2022}
}

@inproceedings{socher2013_sentiment,
  title={Recursive deep models for semantic compositionality over a sentiment treebank},
  author={Socher, Richard and Perelygin, Alex and Wu, Jean and Chuang, Jason and Manning, Christopher D and Ng, Andrew Y and Potts, Christopher},
  booktitle={Proceedings of the 2013 conference on empirical methods in natural language processing},
  pages={1631--1642},
  year={2013}
}

@inproceedings{pang2004_sentiment,
  title={A sentimental education: sentiment analysis using subjectivity summarization based on minimum cuts},
  author={Pang, Bo and Lee, Lillian},
  booktitle={Proceedings of the 42nd Annual Meeting of the Association for Computational Linguistics (ACL-04)},
  pages={271--278},
  year={2004}
}

@inproceedings{rosenthal2017_sentiment,
    title = {{S}em{E}val-2017 Task 4: Sentiment Analysis in {T}witter},
    author = {Rosenthal, Sara and Farra, Noura and Nakov, Preslav},
    booktitle = {Proceedings of the 11th International Workshop on Semantic Evaluation ({S}em{E}val-2017)},
    year = {2017},
    pages = {502--518}
}

@article{scikit_learn,
  title={Scikit-learn: Machine learning in {P}ython},
  author  = {Fabian Pedregosa and Ga{{\"e}}l Varoquaux and Alexandre Gramfort and Vincent Michel and Bertrand Thirion and Olivier Grisel and Mathieu Blondel and Peter Prettenhofer and Ron Weiss and Vincent Dubourg and Jake Vanderplas and Alexandre Passos and David Cournapeau and Matthieu Brucher and Matthieu Perrot and {{\'E}}douard Duchesnay},
  title   = {Scikit-learn: Machine Learning in Python},
  journal={Journal of Machine Learning Research},
  volume={12},
  number  = {85},
  pages={2825--2830},
  year={2011},
  publisher={Journal of Machine Learning Research}
}

@article{liu2019robertarobustlyoptimizedbert_roberta,
 title={{RoBERTa}: A Robustly Optimized {BERT} Pretraining Approach}, 
 author={Yinhan Liu and Myle Ott and Naman Goyal and Jingfei Du and Mandar Joshi and Danqi Chen and Omer Levy and Mike Lewis and Luke Zettlemoyer and Veselin Stoyanov},
 year={2019},
 journal={arXiv preprint arXiv:1907.11692},
}

@inproceedings{yamada-etal-2020-luke,
    title = {{LUKE}: Deep Contextualized Entity Representations with Entity-aware Self-attention},
    author = {Yamada, Ikuya  and
      Asai, Akari  and
      Shindo, Hiroyuki  and
      Takeda, Hideaki  and
      Matsumoto, Yuji},
    booktitle = {Proceedings of the 2020 Conference on Empirical Methods in Natural Language Processing (EMNLP)},
    year = {2020},
    pages = {6442--6454}
}

@article{zaheer2020big,
  title={{Big Bird}: {T}ransformers for longer sequences},
  author={Zaheer, Manzil and Guruganesh, Guru and Dubey, Kumar Avinava and Ainslie, Joshua and Alberti, Chris and Ontanon, Santiago and Pham, Philip and Ravula, Anirudh and Wang, Qifan and Yang, Li and others},
  journal={Advances in neural information processing systems},
  volume={33},
  year={2020}
}

@article{dong2024surveyincontextlearning,
      title={A Survey on In-context Learning}, 
      author={Qingxiu Dong and Lei Li and Damai Dai and Ce Zheng and Jingyuan Ma and Rui Li and Heming Xia and Jingjing Xu and Zhiyong Wu and Tianyu Liu and Baobao Chang and Xu Sun and Lei Li and Zhifang Sui},
      year={2024},
    journal={arXiv preprint arXiv:2301.00234},
}

@article{brown2020language,
  title={Language models are few-shot learners},
  author={Brown, Tom and Mann, Benjamin and Ryder, Nick and Subbiah, Melanie and Kaplan, Jared D and Dhariwal, Prafulla and Neelakantan, Arvind and Shyam, Pranav and Sastry, Girish and Askell, Amanda and others},
  journal={Advances in neural information processing systems},
  volume={33},
  pages={1877--1901},
  year={2020}
}

@article{openai2024gpt4technicalreport,
      title={{GPT-4} Technical Report}, 
      author={OpenAI and Josh Achiam and Steven Adler and Sandhini Agarwal and Lama Ahmad and Ilge Akkaya and Florencia Leoni Aleman and Diogo Almeida and Janko Altenschmidt and Sam Altman and Shyamal Anadkat and Red Avila and Igor Babuschkin and Suchir Balaji and Valerie Balcom and Paul Baltescu and Haiming Bao and Mohammad Bavarian and Jeff Belgum and Irwan Bello and Jake Berdine and Gabriel Bernadett-Shapiro and Christopher Berner and Lenny Bogdonoff and Oleg Boiko and Madelaine Boyd and Anna-Luisa Brakman and Greg Brockman and Tim Brooks and Miles Brundage and Kevin Button and Trevor Cai and Rosie Campbell and Andrew Cann and Brittany Carey and Chelsea Carlson and Rory Carmichael and Brooke Chan and Che Chang and Fotis Chantzis and Derek Chen and Sully Chen and Ruby Chen and Jason Chen and Mark Chen and Ben Chess and Chester Cho and Casey Chu and Hyung Won Chung and Dave Cummings and Jeremiah Currier and Yunxing Dai and Cory Decareaux and Thomas Degry and Noah Deutsch and Damien Deville and Arka Dhar and David Dohan and Steve Dowling and Sheila Dunning and Adrien Ecoffet and Atty Eleti and Tyna Eloundou and David Farhi and Liam Fedus and Niko Felix and Simón Posada Fishman and Juston Forte and Isabella Fulford and Leo Gao and Elie Georges and Christian Gibson and Vik Goel and Tarun Gogineni and Gabriel Goh and Rapha Gontijo-Lopes and Jonathan Gordon and Morgan Grafstein and Scott Gray and Ryan Greene and Joshua Gross and Shixiang Shane Gu and Yufei Guo and Chris Hallacy and Jesse Han and Jeff Harris and Yuchen He and Mike Heaton and Johannes Heidecke and Chris Hesse and Alan Hickey and Wade Hickey and Peter Hoeschele and Brandon Houghton and Kenny Hsu and Shengli Hu and Xin Hu and Joost Huizinga and Shantanu Jain and Shawn Jain and Joanne Jang and Angela Jiang and Roger Jiang and Haozhun Jin and Denny Jin and Shino Jomoto and Billie Jonn and Heewoo Jun and Tomer Kaftan and Łukasz Kaiser and Ali Kamali and Ingmar Kanitscheider and Nitish Shirish Keskar and Tabarak Khan and Logan Kilpatrick and Jong Wook Kim and Christina Kim and Yongjik Kim and Jan Hendrik Kirchner and Jamie Kiros and Matt Knight and Daniel Kokotajlo and Łukasz Kondraciuk and Andrew Kondrich and Aris Konstantinidis and Kyle Kosic and Gretchen Krueger and Vishal Kuo and Michael Lampe and Ikai Lan and Teddy Lee and Jan Leike and Jade Leung and Daniel Levy and Chak Ming Li and Rachel Lim and Molly Lin and Stephanie Lin and Mateusz Litwin and Theresa Lopez and Ryan Lowe and Patricia Lue and Anna Makanju and Kim Malfacini and Sam Manning and Todor Markov and Yaniv Markovski and Bianca Martin and Katie Mayer and Andrew Mayne and Bob McGrew and Scott Mayer McKinney and Christine McLeavey and Paul McMillan and Jake McNeil and David Medina and Aalok Mehta and Jacob Menick and Luke Metz and Andrey Mishchenko and Pamela Mishkin and Vinnie Monaco and Evan Morikawa and Daniel Mossing and Tong Mu and Mira Murati and Oleg Murk and David Mély and Ashvin Nair and Reiichiro Nakano and Rajeev Nayak and Arvind Neelakantan and Richard Ngo and Hyeonwoo Noh and Long Ouyang and Cullen O'Keefe and Jakub Pachocki and Alex Paino and Joe Palermo and Ashley Pantuliano and Giambattista Parascandolo and Joel Parish and Emy Parparita and Alex Passos and Mikhail Pavlov and Andrew Peng and Adam Perelman and Filipe de Avila Belbute Peres and Michael Petrov and Henrique Ponde de Oliveira Pinto and Michael and Pokorny and Michelle Pokrass and Vitchyr H. Pong and Tolly Powell and Alethea Power and Boris Power and Elizabeth Proehl and Raul Puri and Alec Radford and Jack Rae and Aditya Ramesh and Cameron Raymond and Francis Real and Kendra Rimbach and Carl Ross and Bob Rotsted and Henri Roussez and Nick Ryder and Mario Saltarelli and Ted Sanders and Shibani Santurkar and Girish Sastry and Heather Schmidt and David Schnurr and John Schulman and Daniel Selsam and Kyla Sheppard and Toki Sherbakov and Jessica Shieh and Sarah Shoker and Pranav Shyam and Szymon Sidor and Eric Sigler and Maddie Simens and Jordan Sitkin and Katarina Slama and Ian Sohl and Benjamin Sokolowsky and Yang Song and Natalie Staudacher and Felipe Petroski Such and Natalie Summers and Ilya Sutskever and Jie Tang and Nikolas Tezak and Madeleine B. Thompson and Phil Tillet and Amin Tootoonchian and Elizabeth Tseng and Preston Tuggle and Nick Turley and Jerry Tworek and Juan Felipe Cerón Uribe and Andrea Vallone and Arun Vijayvergiya and Chelsea Voss and Carroll Wainwright and Justin Jay Wang and Alvin Wang and Ben Wang and Jonathan Ward and Jason Wei and CJ Weinmann and Akila Welihinda and Peter Welinder and Jiayi Weng and Lilian Weng and Matt Wiethoff and Dave Willner and Clemens Winter and Samuel Wolrich and Hannah Wong and Lauren Workman and Sherwin Wu and Jeff Wu and Michael Wu and Kai Xiao and Tao Xu and Sarah Yoo and Kevin Yu and Qiming Yuan and Wojciech Zaremba and Rowan Zellers and Chong Zhang and Marvin Zhang and Shengjia Zhao and Tianhao Zheng and Juntang Zhuang and William Zhuk and Barret Zoph},
    year={2024},
    journal={arXiv preprint arXiv:2303.08774},
}

@article{qwen3,
 title={{Qwen3} Technical Report}, 
 author={An Yang and Anfeng Li and Baosong Yang and Beichen Zhang and Binyuan Hui and Bo Zheng and Bowen Yu and Chang Gao and Chengen Huang and Chenxu Lv and Chujie Zheng and Dayiheng Liu and Fan Zhou and Fei Huang and Feng Hu and Hao Ge and Haoran Wei and Huan Lin and Jialong Tang and Jian Yang and Jianhong Tu and Jianwei Zhang and Jianxin Yang and Jiaxi Yang and Jing Zhou and Jingren Zhou and Junyang Lin and Kai Dang and Keqin Bao and Kexin Yang and Le Yu and Lianghao Deng and Mei Li and Mingfeng Xue and Mingze Li and Pei Zhang and Peng Wang and Qin Zhu and Rui Men and Ruize Gao and Shixuan Liu and Shuang Luo and Tianhao Li and Tianyi Tang and Wenbiao Yin and Xingzhang Ren and Xinyu Wang and Xinyu Zhang and Xuancheng Ren and Yang Fan and Yang Su and Yichang Zhang and Yinger Zhang and Yu Wan and Yuqiong Liu and Zekun Wang and Zeyu Cui and Zhenru Zhang and Zhipeng Zhou and Zihan Qiu},
 year={2025},
 journal={arXiv preprint arXiv:2505.09388}
}

@article{openai2024o1,
    title={{OpenAI} o1 System Card}, 
    author={OpenAI and Aaron Jaech and Adam Kalai and Adam Lerer and Adam Richardson and Ahmed El-Kishky and Aiden Low and Alec Helyar and Aleksander Madry and Alex Beutel and Alex Carney and Alex Iftimie and Alex Karpenko and Alex Tachard Passos and Alexander Neitz and Alexander Prokofiev and Alexander Wei and Allison Tam and Ally Bennett and Ananya Kumar and Andre Saraiva and Andrea Vallone and Andrew Duberstein and Andrew Kondrich and Andrey Mishchenko and Andy Applebaum and Angela Jiang and Ashvin Nair and Barret Zoph and Behrooz Ghorbani and Ben Rossen and Benjamin Sokolowsky and Boaz Barak and Bob McGrew and Borys Minaiev and Botao Hao and Bowen Baker and Brandon Houghton and Brandon McKinzie and Brydon Eastman and Camillo Lugaresi and Cary Bassin and Cary Hudson and Chak Ming Li and Charles de Bourcy and Chelsea Voss and Chen Shen and Chong Zhang and Chris Koch and Chris Orsinger and Christopher Hesse and Claudia Fischer and Clive Chan and Dan Roberts and Daniel Kappler and Daniel Levy and Daniel Selsam and David Dohan and David Farhi and David Mely and David Robinson and Dimitris Tsipras and Doug Li and Dragos Oprica and Eben Freeman and Eddie Zhang and Edmund Wong and Elizabeth Proehl and Enoch Cheung and Eric Mitchell and Eric Wallace and Erik Ritter and Evan Mays and Fan Wang and Felipe Petroski Such and Filippo Raso and Florencia Leoni and Foivos Tsimpourlas and Francis Song and Fred von Lohmann and Freddie Sulit and Geoff Salmon and Giambattista Parascandolo and Gildas Chabot and Grace Zhao and Greg Brockman and Guillaume Leclerc and Hadi Salman and Haiming Bao and Hao Sheng and Hart Andrin and Hessam Bagherinezhad and Hongyu Ren and Hunter Lightman and Hyung Won Chung and Ian Kivlichan and Ian O'Connell and Ian Osband and Ignasi Clavera Gilaberte and Ilge Akkaya and Ilya Kostrikov and Ilya Sutskever and Irina Kofman and Jakub Pachocki and James Lennon and Jason Wei and Jean Harb and Jerry Twore and Jiacheng Feng and Jiahui Yu and Jiayi Weng and Jie Tang and Jieqi Yu and Joaquin Quiñonero Candela and Joe Palermo and Joel Parish and Johannes Heidecke and John Hallman and John Rizzo and Jonathan Gordon and Jonathan Uesato and Jonathan Ward and Joost Huizinga and Julie Wang and Kai Chen and Kai Xiao and Karan Singhal and Karina Nguyen and Karl Cobbe and Katy Shi and Kayla Wood and Kendra Rimbach and Keren Gu-Lemberg and Kevin Liu and Kevin Lu and Kevin Stone and Kevin Yu and Lama Ahmad and Lauren Yang and Leo Liu and Leon Maksin and Leyton Ho and Liam Fedus and Lilian Weng and Linden Li and Lindsay McCallum and Lindsey Held and Lorenz Kuhn and Lukas Kondraciuk and Lukasz Kaiser and Luke Metz and Madelaine Boyd and Maja Trebacz and Manas Joglekar and Mark Chen and Marko Tintor and Mason Meyer and Matt Jones and Matt Kaufer and Max Schwarzer and Meghan Shah and Mehmet Yatbaz and Melody Y. Guan and Mengyuan Xu and Mengyuan Yan and Mia Glaese and Mianna Chen and Michael Lampe and Michael Malek and Michele Wang and Michelle Fradin and Mike McClay and Mikhail Pavlov and Miles Wang and Mingxuan Wang and Mira Murati and Mo Bavarian and Mostafa Rohaninejad and Nat McAleese and Neil Chowdhury and Neil Chowdhury and Nick Ryder and Nikolas Tezak and Noam Brown and Ofir Nachum and Oleg Boiko and Oleg Murk and Olivia Watkins and Patrick Chao and Paul Ashbourne and Pavel Izmailov and Peter Zhokhov and Rachel Dias and Rahul Arora and Randall Lin and Rapha Gontijo Lopes and Raz Gaon and Reah Miyara and Reimar Leike and Renny Hwang and Rhythm Garg and Robin Brown and Roshan James and Rui Shu and Ryan Cheu and Ryan Greene and Saachi Jain and Sam Altman and Sam Toizer and Sam Toyer and Samuel Miserendino and Sandhini Agarwal and Santiago Hernandez and Sasha Baker and Scott McKinney and Scottie Yan and Shengjia Zhao and Shengli Hu and Shibani Santurkar and Shraman Ray Chaudhuri and Shuyuan Zhang and Siyuan Fu and Spencer Papay and Steph Lin and Suchir Balaji and Suvansh Sanjeev and Szymon Sidor and Tal Broda and Aidan Clark and Tao Wang and Taylor Gordon and Ted Sanders and Tejal Patwardhan and Thibault Sottiaux and Thomas Degry and Thomas Dimson and Tianhao Zheng and Timur Garipov and Tom Stasi and Trapit Bansal and Trevor Creech and Troy Peterson and Tyna Eloundou and Valerie Qi and Vineet Kosaraju and Vinnie Monaco and Vitchyr Pong and Vlad Fomenko and Weiyi Zheng and Wenda Zhou and Wes McCabe and Wojciech Zaremba and Yann Dubois and Yinghai Lu and Yining Chen and Young Cha and Yu Bai and Yuchen He and Yuchen Zhang and Yunyun Wang and Zheng Shao and Zhuohan Li},
    year={2024},
    journal={arXiv preprint arXiv:2412.16720},
}

@article{deepseekai2025deepseekr1,
    title={{DeepSeek-R1}: Incentivizing Reasoning Capability in LLMs via Reinforcement Learning}, 
    author={DeepSeek-AI and Daya Guo and Dejian Yang and Haowei Zhang and Junxiao Song and Ruoyu Zhang and Runxin Xu and Qihao Zhu and Shirong Ma and Peiyi Wang and Xiao Bi and Xiaokang Zhang and Xingkai Yu and Yu Wu and Z. F. Wu and Zhibin Gou and Zhihong Shao and Zhuoshu Li and Ziyi Gao and Aixin Liu and Bing Xue and Bingxuan Wang and Bochao Wu and Bei Feng and Chengda Lu and Chenggang Zhao and Chengqi Deng and Chenyu Zhang and Chong Ruan and Damai Dai and Deli Chen and Dongjie Ji and Erhang Li and Fangyun Lin and Fucong Dai and Fuli Luo and Guangbo Hao and Guanting Chen and Guowei Li and H. Zhang and Han Bao and Hanwei Xu and Haocheng Wang and Honghui Ding and Huajian Xin and Huazuo Gao and Hui Qu and Hui Li and Jianzhong Guo and Jiashi Li and Jiawei Wang and Jingchang Chen and Jingyang Yuan and Junjie Qiu and Junlong Li and J. L. Cai and Jiaqi Ni and Jian Liang and Jin Chen and Kai Dong and Kai Hu and Kaige Gao and Kang Guan and Kexin Huang and Kuai Yu and Lean Wang and Lecong Zhang and Liang Zhao and Litong Wang and Liyue Zhang and Lei Xu and Leyi Xia and Mingchuan Zhang and Minghua Zhang and Minghui Tang and Meng Li and Miaojun Wang and Mingming Li and Ning Tian and Panpan Huang and Peng Zhang and Qiancheng Wang and Qinyu Chen and Qiushi Du and Ruiqi Ge and Ruisong Zhang and Ruizhe Pan and Runji Wang and R. J. Chen and R. L. Jin and Ruyi Chen and Shanghao Lu and Shangyan Zhou and Shanhuang Chen and Shengfeng Ye and Shiyu Wang and Shuiping Yu and Shunfeng Zhou and Shuting Pan and S. S. Li and Shuang Zhou and Shaoqing Wu and Shengfeng Ye and Tao Yun and Tian Pei and Tianyu Sun and T. Wang and Wangding Zeng and Wanjia Zhao and Wen Liu and Wenfeng Liang and Wenjun Gao and Wenqin Yu and Wentao Zhang and W. L. Xiao and Wei An and Xiaodong Liu and Xiaohan Wang and Xiaokang Chen and Xiaotao Nie and Xin Cheng and Xin Liu and Xin Xie and Xingchao Liu and Xinyu Yang and Xinyuan Li and Xuecheng Su and Xuheng Lin and X. Q. Li and Xiangyue Jin and Xiaojin Shen and Xiaosha Chen and Xiaowen Sun and Xiaoxiang Wang and Xinnan Song and Xinyi Zhou and Xianzu Wang and Xinxia Shan and Y. K. Li and Y. Q. Wang and Y. X. Wei and Yang Zhang and Yanhong Xu and Yao Li and Yao Zhao and Yaofeng Sun and Yaohui Wang and Yi Yu and Yichao Zhang and Yifan Shi and Yiliang Xiong and Ying He and Yishi Piao and Yisong Wang and Yixuan Tan and Yiyang Ma and Yiyuan Liu and Yongqiang Guo and Yuan Ou and Yuduan Wang and Yue Gong and Yuheng Zou and Yujia He and Yunfan Xiong and Yuxiang Luo and Yuxiang You and Yuxuan Liu and Yuyang Zhou and Y. X. Zhu and Yanhong Xu and Yanping Huang and Yaohui Li and Yi Zheng and Yuchen Zhu and Yunxian Ma and Ying Tang and Yukun Zha and Yuting Yan and Z. Z. Ren and Zehui Ren and Zhangli Sha and Zhe Fu and Zhean Xu and Zhenda Xie and Zhengyan Zhang and Zhewen Hao and Zhicheng Ma and Zhigang Yan and Zhiyu Wu and Zihui Gu and Zijia Zhu and Zijun Liu and Zilin Li and Ziwei Xie and Ziyang Song and Zizheng Pan and Zhen Huang and Zhipeng Xu and Zhongyu Zhang and Zhen Zhang},
    year={2025},
    journal={arXiv preprint arXiv:2501.12948},
}

@article{breiman2001random,
  title={Random forests},
  author={Breiman, Leo},
  journal={Machine learning},
  volume={45},
  number={1},
  pages={5--32},
  year={2001},
  publisher={Springer}
}

@book{vapnik1995nature_SVM_foundation,
  title={The nature of statistical learning theory},
  author={Vapnik, Vladimir N},
  year={1995},
  publisher={Springer}
}

@article{loshchilov2017_adam,
  title={Decoupled weight decay regularization},
  author={Loshchilov, Ilya and Hutter, Frank},
  journal={arXiv preprint arXiv:1711.05101},
  year={2017}
}

@inproceedings{verhagen2007_tempExp,
  title={{SemEval}-2007 task 15: {T}empeval temporal relation identification},
  author={Verhagen, Marc and Gaizauskas, Robert and Schilder, Frank and Hepple, Mark and Katz, Graham and Pustejovsky, James},
  booktitle={Proceedings of the Fourth International Workshop on Semantic Evaluations},
  pages={75--80},
  year={2007}
}

@inproceedings{verhagen2010_tempExp,
  title={{SemEval}-2010 task 13: {TempEval}-2},
  author={Verhagen, Marc and Sauri, Roser and Caselli, Tommaso and Pustejovsky, James},
  booktitle={Proceedings of the 5th International Workshop on
Semantic Evaluation},
  pages={57--62},
  year={2010}
}

@inproceedings{uzzaman2013_tempExp,
  title={{SemEval}-2013 task 1: {TempEval}-3: Evaluating time expressions, events, and temporal relations},
  author={UzZaman, Naushad and Llorens, Hector and Derczynski, Leon and Allen, James and Verhagen, Marc and Pustejovsky, James},
  booktitle={Proceedings of the Seventh International Workshop on Semantic Evaluation},
  pages={1--9},
  year={2013}
}

@inproceedings{pustejovsky2003_tempExp,
  title={{TimeML}: Robust specification of event and temporal expressions in text.},
  author={Pustejovsky, James and Castano, Jos{\'e} M and Ingria, Robert and Sauri, Roser and Gaizauskas, Robert J and Setzer, Andrea and Katz, Graham and Radev, Dragomir R},
  journal={Proceedings of the 2003 AAAI Spring Symposium},
  year={2003}
}

@article{asahara2014_tempExp,
  title={{BCCWJ-TimeBank}: {T}emporal and event information Annotation on Japanese text},
  author={Asahara, Masayuki and Kato, Sachi and Konishi, Hikari and Imada, Mizuho and Maekawa, Kikuo},
  journal={International Journal of Computational Linguistics {\&} {C}hinese Language Processing},
  year={2014},
  volume={19},
  number={3},
  pages={1--24}
}

\end{document}